\documentclass[11pt,a4paper]{article}
\usepackage[hyperref]{emnlp2018}
\usepackage{xcolor}
\usepackage{times}
\usepackage{latexsym}
\usepackage{amsmath}
\usepackage{url}
\usepackage{algorithm}
\usepackage{amsfonts}
\usepackage[noend]{algpseudocode}
\usepackage{lipsum}
\usepackage{multirow}
\usepackage{framed}
\usepackage{graphicx}
\usepackage{longtable}
\usepackage{tikz}
\usepackage{booktabs}
\usepackage[all]{nowidow}

\graphicspath{{images/},{auto_fig/}}

\aclfinalcopy

\DeclareMathOperator*{\argmax}{\arg\!\max}

\definecolor{color0}{HTML}{FFF5F0}
\definecolor{color1}{HTML}{FEE5D8}
\definecolor{color2}{HTML}{FDCAB5}
\definecolor{color3}{HTML}{FCAB8F}
\definecolor{color4}{HTML}{FC8A6A}
\definecolor{color5}{HTML}{FB694A}
\definecolor{color6}{HTML}{F14432}
\definecolor{color7}{HTML}{D92523}
\definecolor{color8}{HTML}{BC141A}
\definecolor{color9}{HTML}{980C13}
\newcommand*{\mybox}[2]{\tikz[anchor=base,baseline=0pt,rounded corners=0pt, inner sep=0.2mm] \node[fill=#1!60!white] (X) {#2};}

\definecolor{coloranswer}{HTML}{0596FF}
\definecolor{lightblue}{HTML}{3CC7EA}
\definecolor{boxcolor}{HTML}{EEEEEE}
\definecolor{periwinkle}{rgb}{0.8, 0.8, 1.0}
\definecolor{colorsquad}{rgb}{0,1,0}
\definecolor{colorsnli}{rgb}{1,0,0}
\definecolor{colorvqa}{rgb}{1,1,0}

\newcommand{\g}{\, | \,}
\newcommand{\abr}[1]{\textsc{#1}}
\newcommand{\squad}{\textsc{SQuAD}}
\newcommand{\snli}{\textsc{SNLI}}
\newcommand{\vqa}{\textsc{VQA}}
\newcommand{\nlp}{\textsc{nlp}}
\newcommand{\mb}[1]{\boldsymbol{\mathbf{#1}}}
\newcommand{\loo}{leave-one-out}

\newcommand{\jbgcomment}[1]{}
\newcommand{\fscomment}[1]{}
\newcommand{\ewcomment}[1]{}
\newcommand{\prcomment}[1]{}
\newcommand{\alvincomment}[1]{}
\newcommand{\mnicomment}[1]{}
\newcommand{\halcomment}[1]{}

\makeatletter
\def\BState{\State\hskip-\ALG@thistlm}
\makeatother

\title{Pathologies of Neural Models Make Interpretations Difficult}

\author{
  Shi Feng$^1$ Eric Wallace$^1$ Alvin Grissom~II$^2$ Mohit Iyyer$^{3,4}$\\
  {\bf Pedro Rodriguez$^1$ Jordan Boyd-Graber$^1$}\\
  $^1$University of Maryland $^2$Ursinus College\\
  $^3$UMass Amherst $^4$Allen Institute for Artificial Intelligence\\
  {\tt \{shifeng,ewallac2,entilzha,jbg\}@umiacs.umd.edu}, \\
  {\tt agrissom@ursinus.edu}, {\tt miyyer@cs.umass.edu}\\
}

\def\aclpaperid{2051}
\date{}

\begin{document}
\maketitle
\begin{abstract}
One way to interpret neural model predictions is to highlight the
most important input features---for example, a heatmap visualization
over the words in an input sentence. 
In existing interpretation methods for \nlp{}, a word's importance is
determined by either input perturbation---measuring the decrease in
model confidence when that word is removed---or by the gradient with
respect to that word. 
To understand the limitations of these methods,
we use input reduction, which iteratively removes the least important
word from the input. This exposes pathological behaviors of neural models: the
remaining words appear nonsensical to humans and are not the ones determined as
important by interpretation methods. As we confirm with human experiments, the
reduced examples lack information to support the prediction of any label, but
models still make the same predictions with high confidence.  To explain these
counterintuitive results, we draw connections to adversarial examples and
confidence calibration: pathological behaviors reveal difficulties in
interpreting neural models trained with maximum likelihood.  To
mitigate their deficiencies, we fine-tune the models by encouraging high entropy
outputs on reduced examples.  Fine-tuned models become more interpretable under
input reduction without accuracy loss on regular examples.

\end{abstract}

\section{Introduction}
\label{sec:intro}

\begin{figure}[t]
\small
\tikz\node[fill=white!90!colorsquad,inner sep=1pt,rounded corners=0.3cm]{
\begin{tabular}{lp{0.7\columnwidth}}
\textbf{\squad{}} \\
    Context & In 1899, John Jacob Astor IV invested \$100,000 for Tesla to further
develop and produce a new lighting system. Instead, Tesla used the
money to fund his \mybox{coloranswer}{Colorado Springs experiments}. \\\\
Original & What did Tesla spend Astor's money on ? \\
Reduced & did  \\
Confidence & 0.78 $\to$ 0.91 \\
\end{tabular}
}; 
\caption{\squad{} example from the validation set. Given the original
    \emph{Context}, the model makes the same correct prediction (``Colorado
    Springs experiments'') on the \emph{Reduced} question as the
    \emph{Original}, with even higher confidence. For humans, the reduced
    question, ``did'', is nonsensical.}
\label{fig:intro_example}
\end{figure}

Many interpretation methods for neural networks explain the model's prediction
as a counterfactual: how does the prediction
change when the input is modified?  Adversarial
examples~\cite{szegedy2013intriguing,goodfellow2014explaining} highlight the
instability of neural network predictions by showing how small 
perturbations to the input dramatically change the output.

A common, non-adversarial form of model interpretation is feature attribution:
features that are crucial for predictions are highlighted in a heatmap. One can
measure a feature's importance by input perturbation. Given an input for text
classification, a word's importance can be measured by the difference in model
confidence before and after that word is removed from the input---the word is
important if confidence decreases significantly. This is the \loo{}
method~\cite{li2016understanding}.
Gradients can also measure feature importance; for example, a feature
is influential to the prediction if its gradient is a large
positive value. Both perturbation and gradient-based methods can
generate heatmaps, implying that the model's prediction is highly
influenced by the highlighted, important words.

Instead, we study how the model's prediction is influenced by the
\emph{unimportant} words.  We use \textbf{input reduction}, a process that
iteratively removes the unimportant words from the input while maintaining the
model's prediction.  Intuitively, the words remaining after input reduction should
be important for prediction.  Moreover, the words should match the \loo{}
method's selections, which closely align with human
perception~\cite{li2016understanding,murdoch2018cd}.  However, rather than
providing explanations of the original prediction, our reduced examples more
closely resemble adversarial examples. In Figure~\ref{fig:intro_example}, the
reduced input is meaningless to a human but retains the same model prediction
with high confidence. Gradient-based input reduction exposes pathological model
behaviors that contradict what one expects based on existing interpretation
methods.

In Section~\ref{sec:method}, we construct more of these counterintuitive
examples by augmenting input reduction with beam search and experiment with
three tasks: \squad~\cite{rajpurkar2016squad} for reading comprehension,
\snli~\cite{bowman2015snli} for textual entailment, and
\vqa~\cite{antol2015vqa} for visual question answering. 
Input reduction with beam search consistently reduces
the input sentence to very short lengths---often only one or two words---without
lowering model confidence on its original prediction.  The reduced examples
appear nonsensical to humans, which we verify with crowdsourced experiments. In
Section~\ref{sec:confidence}, we draw connections to adversarial
examples and confidence calibration; we explain why the observed pathologies
are a consequence of the overconfidence of neural models. This elucidates
limitations of interpretation methods that rely on model confidence. In
Section~\ref{sec:entropy_training}, we encourage high model uncertainty on
reduced examples with entropy regularization. The pathological model behavior
under input reduction is mitigated, leading to more reasonable reduced examples.

\section{Input Reduction}
\label{sec:method}

To explain model predictions using a set of important words, we must first
define importance. After defining input perturbation and gradient-based
approximation, we describe input reduction with these importance
metrics. Input reduction 
drastically shortens inputs without causing the model to change its
prediction or significantly decrease its confidence.
Crowdsourced experiments confirm that
reduced examples appear nonsensical to humans: input reduction uncovers
pathological model behaviors.

\subsection{Importance from Input Gradient} 

\citet{ribeiro2016lime} and \citet{li2016understanding} define importance by
seeing how confidence changes when a feature is removed; a natural
approximation is to use the gradient~\cite{baehrens2010explain,
simonyan2013deep}.  We formally define these importance metrics in natural
language contexts and introduce the efficient gradient-based approximation.
For each word
in an input sentence, we measure its importance by the change in the
confidence of the original prediction when we remove that word from the sentence.
We switch the sign so that when the confidence decreases, the importance
value is positive.

Formally,
let $\mb{x}=\langle x_1, x_2, \ldots x_n \rangle$ denote the input sentence,
$f(y \g \mb{x})$ the predicted probability of label $y$, and 
$y=\argmax_{y'} f(y' \g \mb{x})$ the original predicted label. The importance is
then
\begin{equation} \label{eq:importance}
    g(x_i\mid \mb{x}) = f(y \g \mb{x}) - f(y \g \mb{x}_{-i}).
\end{equation}
To calculate the importance of each word in a sentence with $n$ words,
we need $n$ forward passes of the model, each time with
one of the words left out.  This is highly inefficient, especially for
longer sentences. Instead, we approximate the importance value with the input
gradient. For each word in the sentence, we calculate the dot product of its
word embedding and the gradient of the output with respect to the embedding. The
importance of $n$ words can thus be computed with a single forward-backward
pass. This gradient approximation has been used for various interpretation
methods for natural language classification models~\cite{li2016visualizing,
arras2016explaining}; see \citet{ebrahimi2017hotflip} for further details on the
derivation. We use this approximation in all our experiments as it selects the
same words for removal as an exhaustive search (no approximation). 

\subsection{Removing Unimportant Words}

Instead of looking at the words with high importance values---what
interpretation methods commonly do---we take a complementary approach and study
how the model behaves when the supposedly unimportant words are removed.  
Intuitively, the important words should remain after the unimportant ones are
removed.

Our input reduction process iteratively removes the unimportant words. At each step,
we remove the word with the lowest importance value until the model changes its
prediction.
We experiment with three popular datasets:
\squad~\cite{rajpurkar2016squad} for reading comprehension,
\snli~\cite{bowman2015snli} for textual entailment,
and \vqa~\cite{antol2015vqa} for visual question answering. We describe each of
these tasks and the model we use below, providing full details in the
Supplement.

In \squad, each example is a context paragraph and a question. The task is to
predict a span in the paragraph as the answer.  We reduce only the question
while keeping the context paragraph unchanged. The model we use is the
\abr{DrQA} Document Reader~\cite{chen2017drqa}.

In \snli, each example consists of two sentences: a premise and a hypothesis.
The task is to predict one of three relationships: entailment, neutral, or
contradiction. We reduce only the hypothesis while keeping the premise
unchanged.  The model we use is Bilateral Multi-Perspective
Matching (\abr{BiMPM})~\cite{wang2017bilateral}.

In \vqa, each example consists of an image and a natural language question.
We reduce only the question while keeping the image unchanged. The model we use
is Show, Ask, Attend, and Answer~\cite{kazemi2017show}.

During the iterative reduction process, we ensure that the prediction does not
change (exact same span for \squad{}); consequently, the model accuracy
on the reduced examples is identical to the original. The predicted label is
used for input reduction and the ground-truth is never revealed. We use the
validation set for all three tasks.

\begin{figure}[t]
\small

\tikz\node[fill=white!90!colorsnli,inner sep=1pt,rounded corners=0.3cm]{
\begin{tabular}{lp{0.7\columnwidth}}
\textbf{\snli} & \\
Premise & Well dressed man and woman dancing in the street \\
Original & Two man is dancing on the street \\
Reduced & dancing \\
Answer & Contradiction \\
Confidence & 0.977 $\to$ 0.706 \\
\end{tabular}
}; 

\tikz\node[fill=white!90!colorvqa,inner sep=1pt,rounded corners=0.3cm]{
\begin{tabular}{lp{0.7\columnwidth}}
\textbf{\vqa} & \\
& \includegraphics[height=1in]{vqa_example} \\
Original & What color is the flower ? \\
Reduced & flower ? \\
Answer & yellow \\
Confidence & 0.827 $\to$ 0.819  \\
\end{tabular}
}; 
\caption{Examples of original and reduced inputs where the models predict the
    same \emph{Answer}. \emph{Reduced} shows the input after reduction.  We
    remove words from the hypothesis for \snli{}, questions for \squad{} and
    \vqa{}.  Given the nonsensical reduced inputs, humans would not be able to
    provide the answer with high confidence, yet, the neural models do.}
\label{fig:reduced_examples}
\end{figure}

Most reduced inputs are nonsensical to humans
(Figure~\ref{fig:reduced_examples}) as they lack information for \emph{any}
reasonable human prediction. However, models make confident predictions, at
times even more confident than the original.

To find the shortest possible reduced inputs (potentially the most meaningless),
we relax the requirement of removing only the least important word and
augment input reduction with beam search. We limit the removal to
the $k$\ \ least important words, where $k$ is the beam size, and decrease the beam
size as the remaining input is shortened.\footnote{We set beam size to $\max(1,
\min(k, L - 3))$ where $k$ is maximum beam size and $L$ is the current length of
the input sentence.} We empirically select beam size five as it produces
comparable results to larger beam sizes with reasonable computation cost. The
requirement of maintaining model prediction is unchanged.
  
\begin{figure*}[t]
    \centering
    \includegraphics[width=.8\textwidth]{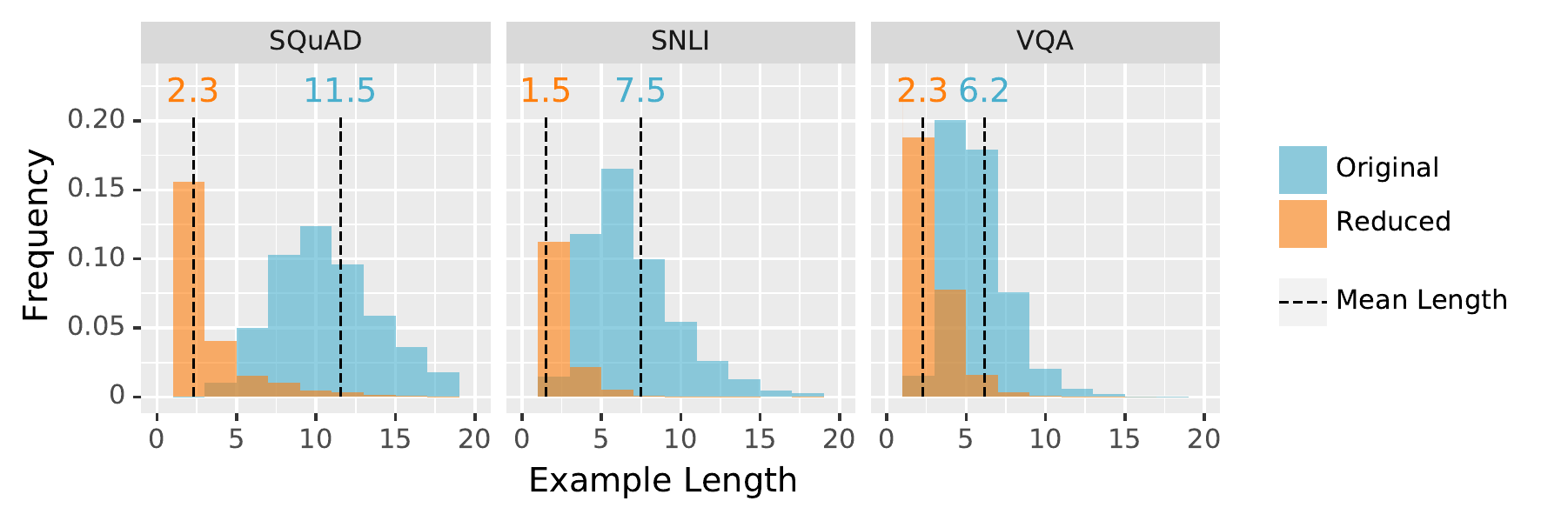}
    \caption{Distribution of input sentence length before and after reduction.
    For all three tasks, the input is often reduced to one or two words without
    changing the model's prediction.}
    \label{fig:lengths_histogram}
\end{figure*}

\begin{figure}[t]
    \centering
    \includegraphics[width=0.5\textwidth]{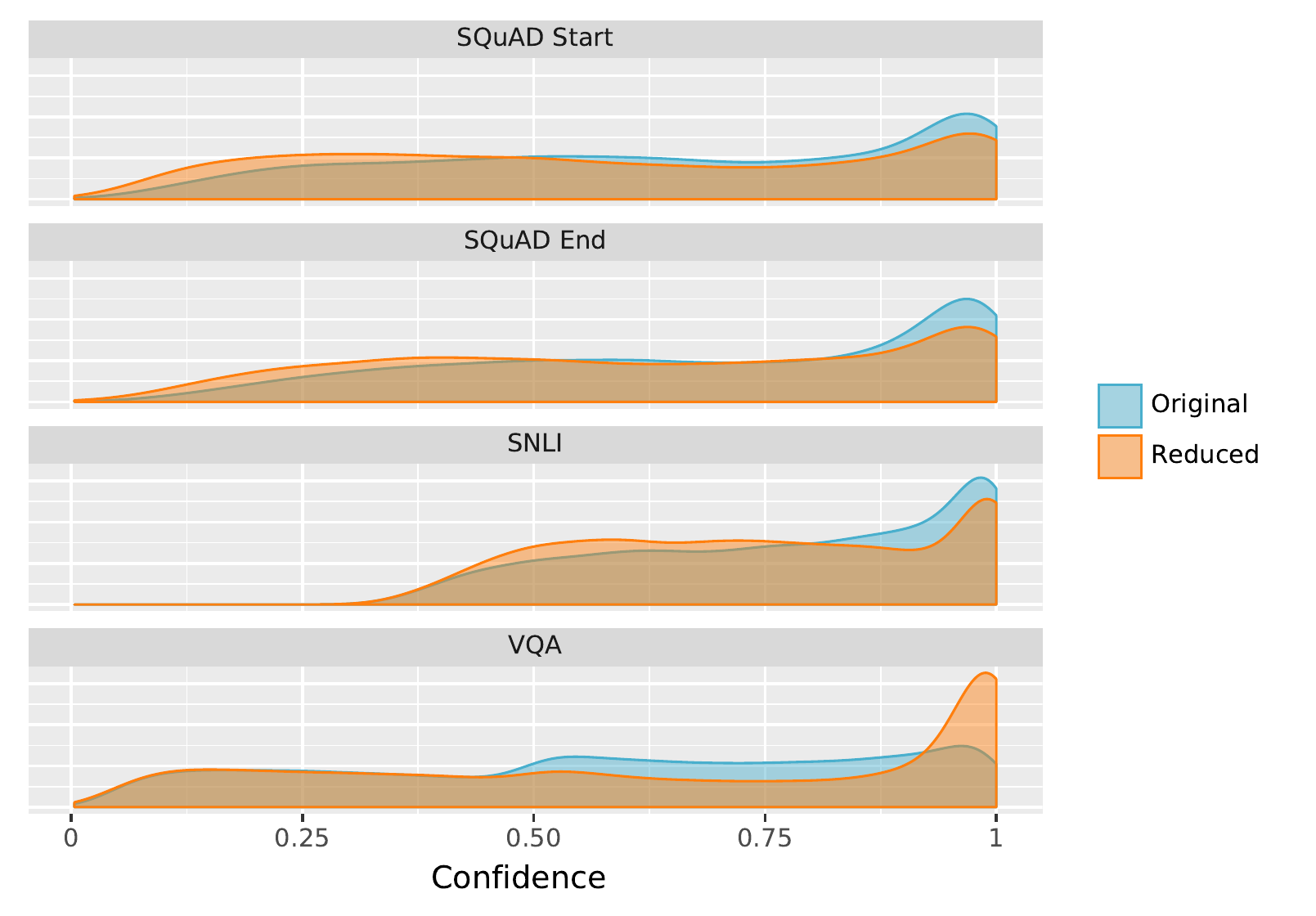}
\caption{Density distribution of model confidence on reduced inputs
    is similar to the original confidence.  In \squad, we predict
    the beginning and the end of the answer span, so we show the
    confidence for both.}
    \label{fig:confidence_density}
\end{figure}

With beam search, input reduction finds extremely short reduced examples with
little to no decrease in the model's confidence on its original
predictions.  Figure~\ref{fig:lengths_histogram} compares the length of input
sentences before and after the reduction. For all three tasks, we can often
reduce the sentence to only one word. Figure~\ref{fig:confidence_density}
compares the model's confidence on original and reduced inputs.  On \squad{} and
\snli{} the confidence decreases slightly, and on \vqa{} the confidence even
increases.

\begin{table}[t]
\centering
\begin{tabular}{lccc}
Dataset & Original & Reduced & vs.\ Random \\
\midrule
\squad  & 80.58 & 31.72 & 53.70 \\
\snli-E & 76.40 & 27.66 & 42.31 \\
\snli-N & 55.40 & 52.66 & 50.64 \\
\snli-C & 76.20 & 60.60 & 49.87 \\
\vqa    & 76.11 & 40.60 & 61.60 \\
\end{tabular}
\caption{Human accuracy on \emph{Reduced} examples drops
    significantly compared to the \emph{Original} examples, however, model
    predictions are identical. The reduced examples also appear random to
    humans---they do not prefer them over random inputs
    (\emph{vs.\ Random}).  For \squad, accuracy is reported using F1
    scores, other numbers are percentages. For
    \snli, we report results on the three classes separately: entailment
    (\emph{-E}), neutral (\emph{-N}), and contradiction (\emph{-C}).}
\label{table:human_results}
\end{table}

\subsection{Humans Confused by Reduced Inputs}
\label{sec:human}

On the reduced examples, the models retain their original predictions
despite short input lengths. The following experiments examine whether
these predictions are justified or pathological, based on how humans react to
the reduced inputs.

For each task, we sample 200 examples that are correctly classified by
the model and generate their reduced examples. In the first setting, we
compare the human accuracy on original and reduced examples.
We recruit two groups of crowd workers and task them with textual entailment,
reading comprehension, or visual question answering. We show one group the
original inputs and the other the reduced.  Humans are no longer able to give
the correct answer, showing a significant accuracy loss on all three tasks
(compare \emph{Original} and \emph{Reduced} in Table~\ref{table:human_results}).

The second setting examines how random the reduced examples appear to
humans. For each of the original examples, we generate a version where words are
randomly removed until the
length matches the one generated by input reduction. We present the original example
along with the two reduced examples and ask crowd workers their preference between the two
reduced ones. The workers' choice is almost fifty-fifty (the \emph{vs.\ Random}
in Table~\ref{table:human_results}): the reduced examples appear almost
random to humans.

These results leave us with two puzzles: why are the models highly confident on the
nonsensical reduced examples? And why, when the \loo{} method selects important
words that appear reasonable to humans, the input reduction process selects ones
that are nonsensical?

\section{Making Sense of Reduced Inputs}
\label{sec:confidence}

Having established the incongruity of our definition of importance
\textit{vis-\`a-vis}
human judgements, we now investigate possible explanations for these results.
We explain why model confidence can empower methods such as \loo{} to
generate reasonable interpretations but also lead to
pathologies under input reduction. We attribute these results to two
issues of neural models.

\subsection{Model Overconfidence}
\label{sec:overconfidence}

Neural models are overconfident in their predictions~\cite{guo2017calibration}.
One explanation for overconfidence is overfitting: the model overfits the
negative log-likelihood loss during training by learning to output low-entropy
distributions over classes.  Neural models are also overconfident on examples
outside the training data distribution.  As \citet{goodfellow2014explaining}
observe for image classification, samples from pure noise can sometimes
trigger highly confident predictions. These so-called \emph{rubbish examples}
are degenerate inputs that a human would trivially classify as not belonging to
any class but for which the model predicts with high confidence.
\citet{goodfellow2014explaining} argue that the rubbish examples exist for the
same reason that adversarial examples do: the surprising linear nature of neural
models. In short, the confidence of a neural model is not a
robust estimate of its prediction uncertainty.

Our reduced inputs satisfy the definition of rubbish examples: humans have a
hard time making predictions based on the reduced inputs
(Table~\ref{table:human_results}), but models make predictions with high
confidence (Figure~\ref{fig:confidence_density}). Starting from a valid example,
input reduction transforms it into a rubbish example.

The nonsensical, almost random results are best explained by looking at a
complete reduction path (Figure~\ref{fig:path}). In this example, the transition
from valid to rubbish happens immediately after the first step: following the removal of
``Broncos'', humans can no longer determine which team the question
is asking about, but model confidence remains high.  Not being able to lower its
confidence on rubbish examples---as it is not trained to do so---the model
neglects ``Broncos'' and eventually the process generates nonsensical
results.

In this example, the \loo{} method will not highlight ``Broncos''. However, this
is not a failure of the interpretation method but of the model itself.  The model
assigns a low importance to ``Broncos'' in the first step,
causing it to be removed---\loo{} would be able to expose this particular issue
by not highlighting ``Broncos''.  However, in cases where a similar issue
only appear after a few unimportant words are removed,
the \loo{} method would fail to expose the unreasonable model behavior.

Input reduction can expose deeper issues of model overconfidence and
stress test a model's uncertainty estimation and interpretability.

\begin{figure}[t]
\scriptsize
\tikz\node[fill=white!90!colorsquad,inner sep=1pt,rounded corners=0.3cm]{
\begin{tabular}{p{.95\columnwidth}}
\textbf{\squad{}}\\
Context: The Panthers used the San Jose State practice facility and stayed at the San
Jose Marriott. The Broncos practiced at
\mybox{coloranswer}{Stanford University}
and stayed at the Santa Clara Marriott.\\\\
Question:\\
(0.90, 0.89) Where did the \underline{Broncos} practice for the Super Bowl ? \\
(0.92, 0.88) Where did \underline{the} practice for the Super Bowl ? \\
(0.91, 0.88) Where did practice \underline{for} the Super Bowl ? \\
(0.92, 0.89) Where did practice the Super \underline{Bowl} ? \\
(0.94, 0.90) Where did practice \underline{the} Super ? \\
(0.93, 0.90) \underline{Where} did practice Super ?\\
(0.40, 0.50) did practice Super ? \\
\end{tabular}
}; 
\caption{A reduction path for a \squad{} validation example. The model
    prediction is always correct and its confidence stays high (shown
    on the left in parentheses) throughout the reduction.  Each line
    shows the input at that step with an \underline{underline}
    indicating the word to remove next. The question becomes
    unanswerable immediately after ``Broncos'' is removed in the first
    step. However, in the context of the original question,
    ``Broncos'' is the least important word according to the input
    gradient.}
\label{fig:path}
\end{figure}

\subsection{Second-order Sensitivity}
\label{sec:secondorder}

\begin{figure}[t]
\scriptsize
\tikz\node[fill=white!90!colorsquad,inner sep=1pt,rounded corners=0.3cm]{
\begin{tabular}{p{.95\columnwidth}}
\textbf{\squad{}} \\
Context: QuickBooks sponsored a ``Small Business Big Game'' contest, in
which Death Wish Coffee had a 30-second commercial aired free of
charge courtesy of QuickBooks. \mybox{coloranswer}{Death Wish
Coffee} beat out nine other contenders from across the United
States for the free advertisement.\\\\
Question:\\
\input{sections/figure_4}
\end{tabular}
}; 
\caption{Heatmap generated with \loo{} shifts drastically despite only
    removing the least important word (\underline{underlined}) at each step.
    For instance, ``advertisement'', is the most important word in step two but
    becomes the least important in step three.}
\label{fig:heatmap}
\end{figure}

So far, we have seen that the output of a neural model is sensitive to small
changes in its input. We call this \emph{first-order}
sensitivity, because interpretation based on input gradient is a
first-order Taylor expansion of the model near the
input~\cite{simonyan2013deep}.
However, the \emph{interpretation} also shifts drastically with small
input changes (Figure~\ref{fig:heatmap}). We call this
\emph{second-order} sensitivity.

The shifting heatmap suggests a mismatch between the model's first- and
second-order sensitivities. The heatmap shifts when, with respect to the removed
word, the model has low first-order sensitivity but high second-order
sensitivity.

Similar issues complicate comparable
interpretation methods for image classification models. For example,
\citet{ghorbani2017interpretation} modify image inputs so the
highlighted features in the interpretation change while maintaining the same
prediction. To achieve this, they iteratively
modify the input to maximize changes in
the distribution of feature importance.
In contrast, the shifting heatmap we observe occurs by only removing the least
impactful features without a
targeted optimization. 
They also speculate that the steepest gradient direction for
the first- and second-order sensitivity values are generally orthogonal. Loosely
speaking, the shifting heatmap suggests that the direction of the smallest
gradient value can sometimes align with very steep changes in second-order
sensitivity.

When explaining individual model predictions, the heatmap
suggests that the prediction is made based on a weighted combination of words, as
in a linear model, which is not true unless the model is indeed taking a
weighted sum such as in a \abr{dan}~\cite{iyyer2015deep}. When the model
composes representations by a non-linear combination of words, a linear
interpretation oblivious to second-order sensitivity can be misleading.

\section{Mitigating Model Pathologies}
\label{sec:entropy_training}

The previous section explains the observed pathologies from the perspective of
overconfidence: models are too certain on rubbish examples when they should not make
\emph{any} prediction. Human experiments in Section~\ref{sec:human}
confirm that the reduced examples fit the definition of rubbish examples. Hence, a
natural way to mitigate the pathologies is to maximize model uncertainty on the
reduced examples.

\subsection{Regularization on Reduced Inputs}

To maximize model uncertainty on reduced examples, we use the entropy of the output
distribution as an objective.  Given a model $f$ trained on a
dataset~$(\mathcal{X}, \mathcal{Y})$, we generate reduced examples
using input reduction
for all training examples $\mathcal{X}$. Beam search often yields multiple
reduced versions with the same minimum length for each input $\mb{x}$, and we
collect all of these versions together to form $\tilde{\mathcal{X}}$ as the
``negative'' example set.

Let $\mathbb{H}\left(\cdot\right)$ denote the entropy and $f(y \g \mb{x})$
denote the probability of the model predicting $y$ given $\mb{x}$.
We fine-tune the existing model to 
simultaneously maximize the log-likelihood on regular examples and the
entropy on reduced examples:
\begin{equation} \label{regularizer}
\sum_{(\mb{x}, y) \in (\mathcal{X}, \mathcal{Y})}\log(f(y \g \mb{x})) \
+ \lambda\sum_{\tilde{\mb{x}}\in \tilde{\mathcal{X}}}
\mathbb{H}\left(f(y \g \tilde{\mb{x}})\right),
\end{equation}
where hyperparameter $\lambda$ controls the trade-off between the two
terms. Similar entropy regularization is used by
\citet{pereyra2017regularizing}, but not in combination with input
reduction; their entropy term is calculated on regular examples rather
than reduced examples.

\subsection{Regularization Mitigates Pathologies}

On regular examples, entropy regularization does no harm to model accuracy, with
a slight increase for \squad{} (\emph{Accuracy} in
Table~\ref{table:tuned_machine}).
    
After entropy regularization, input reduction produces more reasonable
reduced inputs (Figure~\ref{fig:tuned_examples}).
In the \squad{} example from Figure~\ref{fig:intro_example}, the reduced
question changed from ``did'' to ``spend Astor money on ?'' after fine-tuning.
The average length of reduced examples also increases across all tasks
(\emph{Reduced length} in Table~\ref{table:tuned_machine}).
To verify that model overconfidence is indeed mitigated---that the reduced
examples are less ``rubbish'' compared to before fine-tuning---we repeat the
human experiments from Section~\ref{sec:human}.

Human accuracy increases across all three tasks (Table~\ref{table:tuned_human}).
We also repeat the \emph{vs.\ Random} experiment: we re-generate the random
examples to match the lengths of the new reduced examples from input
reduction, and find humans now prefer the reduced examples to random ones.
The increase in both human performance and preference suggests that the reduced
examples are more reasonable; model pathologies have been mitigated.

While these results are promising, it is not clear whether our input reduction
method is necessary to achieve them. To provide a baseline, we fine-tune
models using inputs randomly reduced to the same lengths as the ones generated
by input reduction. This baseline improves neither the model
accuracy on regular examples nor interpretability under input reduction (judged
by lengths of reduced examples).  Input reduction is effective in generating
negative examples to counter model overconfidence.

\begin{figure}[t]
\small
\tikz\node[fill=white!90!colorsquad,inner sep=1pt,rounded corners=0.3cm]{
\begin{tabular}{lp{0.7\columnwidth}}
\textbf{\squad} & \\
Context & In 1899, John Jacob Astor IV invested \$100,000 for Tesla to further 
develop and produce a new lighting system. Instead, Tesla used the
money to fund his \mybox{coloranswer}{Colorado Springs experiments}. \\
Original & What did Tesla spend Astor's money on ? \\
Answer & Colorado Springs experiments \\
Before & did \\
After & spend Astor money on ? \\
Confidence & 0.78 $\to$ 0.91 $\to$ 0.52 \\
\end{tabular}
}; 

\tikz\node[fill=white!90!colorsnli,inner sep=1pt,rounded corners=0.3cm]{
\begin{tabular}{lp{0.7\columnwidth}}
\textbf{\snli} & \\
Premise & Well dressed man and woman dancing in the street \\
Original & Two man is dancing on the street \\
Answer & Contradiction \\
Before & dancing \\
After & two man dancing \\
Confidence & 0.977 $\to$ 0.706 $\to$ 0.717 \\
\end{tabular}
}; 

\tikz\node[fill=white!90!colorvqa,inner sep=1pt,rounded corners=0.3cm]{
\begin{tabular}{lp{0.7\columnwidth}}
\textbf{\vqa} & \\
Original & What color is the flower ? \\
Answer & yellow \\
Before & flower ? \\
After & What color is flower ? \\
Confidence & 0.847 $\to$ 0.918 $\to$ 0.745 \\
\end{tabular}
}; 
\caption{\squad{} example from Figure~\ref{fig:intro_example}, \snli{} and
    \vqa{} (image omitted) examples from Figure~\ref{fig:reduced_examples}. We
    apply input reduction to models both \emph{Before} and \emph{After} entropy
    regularization. The models still predict the same \emph{Answer}, but the
    reduced examples after fine-tuning appear more reasonable to humans.}
\label{fig:tuned_examples}
\end{figure}

\begin{table}[t]
\centering
\begin{tabular}{lcccc}
& \multicolumn{2}{c}{Accuracy} & \multicolumn{2}{c}{Reduced length} \\
\cmidrule(l){2-3} \cmidrule(l){4-5}
& Before & After & Before & After \\\midrule
\squad & 77.41 & 78.03 & 2.27 & 4.97 \\
\snli  & 85.71 & 85.72 & 1.50 & 2.20 \\
\vqa   & 61.61 & 61.54 & 2.30 & 2.87
\end{tabular}
\caption{Model \emph{Accuracy} on regular validation
    examples remains largely unchanged after fine-tuning. However, the length of the reduced
    examples (\emph{Reduced length}) increases on all three tasks, making them
    less likely to appear nonsensical to humans.}
\label{table:tuned_machine}
\end{table}

\begin{table}[t]
\centering
\resizebox{\linewidth}{!}{\begin{tabular}{lcccc}        
& \multicolumn{2}{c}{Accuracy} & \multicolumn{2}{c}{vs.\ Random} \\
\cmidrule(l){2-3} \cmidrule(l){4-5}
& \multicolumn{1}{l}{Before} & \multicolumn{1}{l}{After} & \multicolumn{1}{l}{Before} & \multicolumn{1}{l}{After} \\\midrule
\squad  & 31.72 & 51.61 & 53.70 & 62.75 \\
\snli-E & 27.66 & 32.37 & 42.31 & 50.62 \\
\snli-N & 52.66 & 50.50 & 50.64 & 58.94 \\
\snli-C & 60.60 & 63.90 & 49.87 & 56.92 \\
\vqa    & 40.60 & 51.85 & 61.60 & 61.88 \\
\end{tabular}}
\caption{Human \emph{Accuracy} increases after fine-tuning the
    models. Humans also prefer gradient-based reduced examples over
    randomly reduced ones, indicating that the reduced examples are more
    meaningful to humans after regularization.}
\label{table:tuned_human}
\end{table}

\section{Discussion}
\label{sec:discussion}

Rubbish examples have been studied in the image
domain~\cite{goodfellow2014explaining, nguyen2015fooled}, but to our knowledge not
for \nlp{}.  Our input reduction process gradually transforms a valid input
into a rubbish example. We can often determine which word's removal
causes the transition to occur---for example, removing ``Broncos'' in
Figure~\ref{fig:path}.  These rubbish examples are
particularly interesting, as they are also adversarial:
the difference from a valid example is small, unlike image rubbish examples
generated from pure noise which are far outside the training data
distribution.

The robustness of \nlp{} models has been studied
extensively~\cite{papernot2016crafting, jia2017adversarial,
iyyer2018scpn,ribeiro2018semantically},
and most studies define adversarial examples similar to the image domain: small
perturbations to the input lead to large changes in the output.
HotFlip~\cite{ebrahimi2017hotflip} uses a gradient-based approach, 
similar to image adversarial examples, to flip the model prediction by
perturbing a few characters or words.
Our work and \citet{belinkov2017synthetic} both identify cases where noisy
user inputs become adversarial by accident: common misspellings break neural
machine translation models; we show that incomplete user input can lead to
unreasonably high model confidence.

Other failures of interpretation methods have been explored in the image
domain. The sensitivity issue of gradient-based interpretation methods,
similar to our shifting heatmaps, are observed by
\citet{ghorbani2017interpretation} and \citet{kindermans2017unreliability}. They
show that various forms of input perturbation---from adversarial changes to simple
constant shifts in the image input---cause significant changes in the
interpretation. \citet{ghorbani2017interpretation} make a similar observation
about second-order sensitivity, that ``the fragility of interpretation is
orthogonal to fragility of the prediction''.

Previous work studies biases in the annotation process that lead to
datasets easier than desired or expected which eventually induce pathological
models. We attribute our observed pathologies primarily to the lack of accurate
uncertainty estimates in neural models trained with maximum likelihood.
 \snli{} hypotheses contain artifacts
that allow \emph{training} a model without the
premises~\cite{gururangan2018annotation}; we apply input reduction at
\emph{test} time to the
hypothesis. Similarly, \vqa{} images are surprisingly unimportant for
training a model; we reduce the question. The recent \squad{}
2.0~\cite{rajpurkar2018know} augments the original
reading comprehension task with an uncertainty modeling requirement, the
goal being to make the task more realistic and challenging.

Section~\ref{sec:overconfidence} explains the pathologies from the
overconfidence perspective. One explanation for overconfidence is overfitting: 
\citet{guo2017calibration} show that, late in maximum likelihood training,
the model learns to minimize loss by outputting low-entropy distributions without
improving validation accuracy. To examine if overfitting can explain the input
reduction results, we run input reduction using \abr{DrQA} model checkpoints from
every training epoch. Input reduction still achieves similar results on earlier
checkpoints, suggesting that better convergence in maximum likelihood
training cannot fix the issues by itself---we need new training objectives with
uncertainty estimation in mind.

\subsection{Methods for Mitigating Pathologies}

We use the reduced examples generated by input reduction to regularize the model
and improve its interpretability. This resembles adversarial
training~\cite{goodfellow2014explaining}, where adversarial examples are added to
the training set to improve model robustness. The objectives are different:
entropy regularization
encourages high uncertainty on rubbish examples, while adversarial training
makes the model less sensitive to adversarial perturbations.

\citet{pereyra2017regularizing} apply entropy regularization on regular examples
from the start of training to improve model generalization. A similar
method is label smoothing~\cite{szegedy2016rethinking}.  In comparison, we
fine-tune a model with entropy regularization on the reduced examples
for better uncertainty estimates and interpretations.

To mitigate overconfidence, \citet{guo2017calibration} propose
\emph{post-hoc} fine-tuning a model's confidence with Platt scaling.
This method adjusts the softmax function's temperature parameter using a
small held-out dataset to align confidence with accuracy.
However, because the output is calibrated using the entire
confidence distribution, not individual values, this does not reduce
overconfidence on specific inputs, such as the reduced examples.

\subsection{Generalizability of Findings}

To highlight the erratic model predictions on short examples and
provide a more intuitive demonstration, we present paired-input tasks. On
these tasks, the short lengths of reduced questions and hypotheses obviously
contradict the necessary number of words for a human prediction (further
supported by our human studies).
We also apply input reduction to single-input tasks including
sentiment analysis~\cite{mass2011imdb} and
Quizbowl~\cite{boydgraber2012besting}, achieving similar results.

Interestingly, the reduced examples transfer to other architectures.
In particular, when we feed fifty reduced \snli{} inputs from each
class---generated with the \abr{BiMPM}
model~\cite{wang2017bilateral}---through the Decomposable Attention
Model~\cite{parikh2016decomposable},\footnote{\url{http://demo.allennlp.org/textual-entailment}}
the same prediction is triggered 81.3\% of the time.

\section{Conclusion}
\label{sec:conclusion}
We introduce input reduction, a process that iteratively removes unimportant
words from an input while maintaining a model's prediction. Combined with
gradient-based importance estimates often used for interpretations, we
expose pathological behaviors of neural
models. Without lowering model confidence on its
original prediction, an input sentence can be reduced to the point
where it appears nonsensical, often consisting of one or two words. Human
accuracy degrades when shown the reduced examples instead of the original, in
contrast to neural models which maintain their original predictions.

We explain these pathologies with known issues of neural models:
overconfidence and sensitivity to small input changes. The nonsensical reduced
examples are caused by inaccurate uncertainty estimates---the model is not able
to lower its confidence on inputs that do not belong to any label.
The second-order sensitivity is another issue why
gradient-based interpretation methods may fail to align with human perception:
a small change in the input can cause, at the same time, a minor change in the
prediction but a large change in the interpretation. Input reduction
perturbs the input multiple times and can expose deeper issues of
model overconfidence and oversensitivity that other methods cannot.
Therefore, it can be used to stress test the interpretability of a model.

Finally, we fine-tune the models by maximizing entropy on
reduced examples to mitigate the deficiencies. This improves
interpretability without sacrificing model accuracy on regular examples. 

To properly interpret neural models, it is important to
understand their fundamental
characteristics: the nature of their decision surfaces, robustness
against adversaries, and limitations of their training objectives.
We explain fundamental difficulties of interpretation due to
pathologies in neural models trained with maximum likelihood.  Our
work suggests several future directions to improve interpretability:
more thorough evaluation of interpretation methods, better uncertainty
and confidence estimates, and interpretation beyond bag-of-word heatmap.

\section*{Acknowledgments}
Feng was supported under subcontract to Raytheon BBN Technologies by DARPA award
HR0011-15-C-0113.
JBG is supported by NSF Grant IIS1652666.
Any opinions, findings, conclusions, or recommendations expressed here are those
of the authors and do not necessarily reflect the view of the sponsor.
The authors would like to thank Hal Daum{\'e} III, Alexander M. Rush, Nicolas
Papernot, members of the CLIP lab at the University of Maryland, and the anonymous
reviewers for their feedback.

\bibliography{journal-full,fs,jbg}
\bibliographystyle{emnlp2018}

\clearpage

\appendix
\appendix
\section{Model and Training Details}
\label{sec:model_supplemental}

\paragraph{\squad{}: Reading Comprehension}
The model we use for \squad{} is the Document Reader of
\abr{DrQA}~\cite{chen2017drqa}. We use the open source
implementation from \url{https://github.com/hitvoice/DrQA}. The model
represents the words of the paragraph as the concatenation of: 
\abr{GloVe} embeddings~\cite{pennington2014glove}, other word-level
features, and attention scores between the words in the paragraph and
the question. The model runs a \abr{BiLSTM} over the sequence and
takes the resulting hidden states at each time step as a word's final
vector representation. Two classifiers then predict the beginning and
end of the answer span independently. The span with the highest
product of beginning and end probabilities is selected as the answer.
In input reduction, we keep the exact same span.

\paragraph{\snli{}: Textual Entailment}
The model we use for \snli{} is the Bilateral Multi-Perspective
Matching Model (\abr{BiMPM})~\cite{wang2017bilateral}.
We use the open source implementation from
\url{https://github.com/galsang/BIMPM-pytorch}. The model runs a
\abr{BiLSTM} over the \abr{GloVe} word embeddings of both
the premise and hypothesis. It then computes ``matches'' between the two
sentences at multiple perspectives, aggregates the results using
another \abr{BiLSTM}, and feeds the resulting representation
through fully-connected layers for
classification.

\paragraph{\vqa{}: Visual Question Answering}
The model we use for \vqa{} is Show, Ask, Attend and
Answer~\cite{kazemi2017show}.  We use the open source implementation
from \url{https://github.com/Cyanogenoid/pytorch-vqa}.  The model uses
a ResNet-152~\cite{he2016resnet} to represent the image and an
\abr{LSTM} to represent the natural language question.  The model then
computes attention regions based on the two representations.  All
representations are finally combined to predict the answer.

\paragraph{Entropy Regularized Fine-Tuning}
To optimize the objective of Equation ~\ref{regularizer}, we alternate
updates on the two terms. We update the model on two
batches of regular examples for the first term (maximum likelihood), followed
by two batches of reduced examples for the second term (maximum entropy). The batches of
reduced examples are randomly sampled from the collection of reduced
inputs. We use two separate Adam~\cite{kingma2015adam} optimizers for
the two terms.  For \squad{} and \snli{}, we use a learning rate of
$2e^{-4}$ and $\lambda$ of $1e^{-3}$.  For \vqa{}, we use a learning
rate of $1e^{-4}$ and $\lambda$ of $1e^{-4}$.

\section{More Examples}
\label{sec:example_supplemental}

This section provides more examples of input reduction.  We show
the original and reduced examples, which are generated on the models both 
\emph{Before} and \emph{After} fine-tuning with the entropy regularization from
Section~\ref{sec:entropy_training}. All examples are correctly classified by the model.

\begin{figure*}[t]
\begin{tabular}{lp{0.85\textwidth}}
\textbf{\squad} & \\\hline

Context & In 1899, John Jacob Astor IV invested \$ 100,000 for Tesla
    to further develop and produce a new lighting system. Instead,
    Tesla used the money to fund his \mybox{coloranswer}{Colorado
    Springs experiments}. \\
Original & What did Tesla spend Astor's money on ? (0.78) \\
Before & did \\
After & spend Astor money on ? \\\hline

Context & The Colorado experiments had prepared Tesla for the
    establishment of the \mybox{coloranswer}{trans-Atlantic wireless
    telecommunications facility} known as Wardenclyffe near Shoreham,
    Long Island. \\
Original & What did Tesla establish following his Colorado experiments ? \\
Before & experiments ? \\
After & What Tesla establish experiments ? \\\hline

Context & The Broncos defeated the Pittsburgh Steelers in the
    divisional round, 23–16, by scoring 11 points in the final three
    minutes of the game. They then beat the defending Super Bowl XLIX
    champion \mybox{coloranswer}{New England Patriots} in the AFC
    Championship Game, 20–18, by intercepting a pass on New England's
    2-point conversion attempt with 17 seconds left on the clock.
    Despite Manning's problems with interceptions during the season,
    he didn't throw any in their two playoff games. \\
Original & Who did the Broncos defeat in the AFC Championship game ? \\
Before & Who the defeat the \\
After & Who Broncos defeat AFC Championship game \\\hline

Context & In 2014, economists with the Standard \& Poor's rating
    agency concluded that the widening disparity between the U.S.'s
    wealthiest citizens and the rest of the nation had slowed its
    recovery from the 2008-2009 recession and made it more prone to
    \mybox{coloranswer}{boom-and-bust cycles}. To partially remedy the
    wealth gap and the resulting slow growth, S\&P recommended
    increasing access to education. It estimated that if the average
    United States worker had completed just one more year of school,
    it would add an additional \$105 billion in growth to the
    country's economy over five years. \\
Original & What is the United States at risk for because of the
    recession of 2008 ? \\
Before & is the risk the \\
After & What risk because of the 2008 \\\hline

Context & The Central Region, consisting of present-day Hebei,
    Shandong, Shanxi, the south-eastern part of present-day Inner
    Mongolia and the Henan areas to the north of the Yellow River, was
    considered the most important region of the dynasty and directly
    governed by the Central Secretariat (or Zhongshu Sheng) at
    \mybox{coloranswer}{Khanbaliq} (modern Beijing); similarly,
    another top-level administrative department called the Bureau of
    Buddhist and Tibetan Affairs (or Xuanzheng Yuan) held
    administrative rule over the whole of modern-day Tibet and a part
    of Sichuan, Qinghai and Kashmir. \\
Original & Where was the Central Secretariat based ? \\
Before & the based \\
After & was Central Secretariat based \\\hline

Context & It became clear that managing the Apollo program would
    exceed the capabilities of Robert R. Gilruth's Space Task Group,
    which had been directing the nation's manned space program from
    NASA's Langley Research Center. So Gilruth was given authority to
    grow his organization into a new NASA center, the Manned
    Spacecraft Center (MSC). A site was chosen in
    \mybox{coloranswer}{Houston, Texas}, on land donated by Rice
    University, and Administrator Webb announced the conversion on
    September 19, 1961. It was also clear NASA would soon outgrow its
    practice of controlling missions from its Cape Canaveral Air Force
    Station launch facilities in Florida, so a new Mission Control
    Center would be included in the MSC. \\
Original & Where was the Manned Spacecraft Center located ? \\
Before & Where \\
After & Where was Manned Center located \\\hline

\end{tabular}
\end{figure*}

\begin{figure*}[t]
\centering
\begin{tabular}{lp{0.85\textwidth}}
\textbf{\snli} & \\\hline

Premise & a man in a black shirt is playing a guitar . \\
Original & the man is wearing a blue shirt . \\
Answer & contradiction \\
Before & blue \\
After & man blue shirt \\\hline

Premise & two female basketball teams watch suspenseful as the
    basketball nears the basket . \\
Original & two teams are following the movement of a basketball . \\
Answer & entailment \\
Before & . \\
After & movement \\\hline

Premise & trucks racing \\
Original & there are vehicles . \\
Answer & entailment \\
Before & are \\
After & vehicles \\\hline

Premise & a woman , whose face can only be seen in a mirror , is
    applying eyeliner in a dimly lit room . \\
Original & a man is playing softball . \\
Answer & contradiction \\
Before & man playing \\
After & a man \\\hline

Premise & a tan dog jumping over a red and blue toy \\
Original & a dog playing \\
Answer & entailment \\
Before & dog \\
After & playing \\\hline

Premise & a man in a white shirt has his mouth open and is adjusting dials . \\
Original & a man is sleeping . \\
Answer & contradiction \\
Before & sleeping \\
After & man sleeping \\\hline
\end{tabular}
\end{figure*}

\begin{figure*}
\begin{tabular}{lp{0.85\columnwidth}lp{0.85\columnwidth}}
\textbf{\vqa} \\\hline\\

& \includegraphics[height=1in]{135}     & & \includegraphics[height=1in]{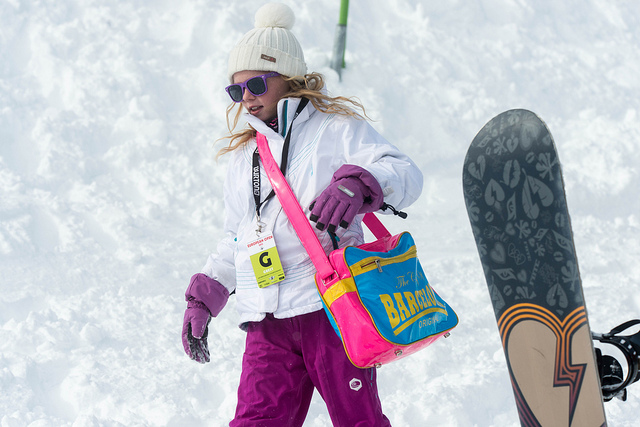}\\
Original & what color is the flower     & Original & what letter is on the lanyard \\
Answer & yellow                         & Answer & g \\
Before & flower                         & Before & letter is on lanyard \\
After & what color is flower            & After & what letter is the lanyard \\\hline\\
                                         
& \includegraphics[height=1in]{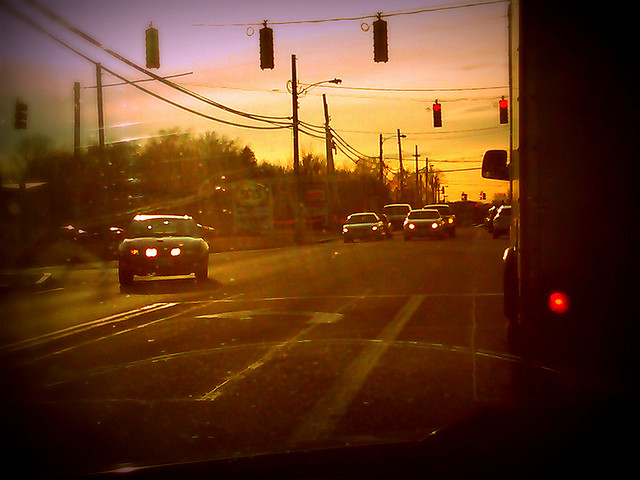}     & & \includegraphics[height=1in]{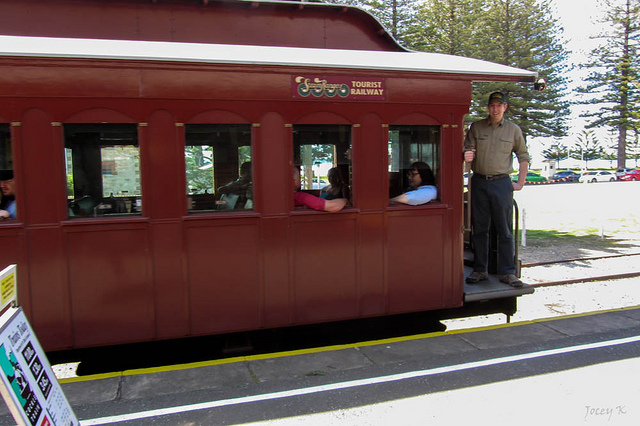}\\
Original & are the cars lights on       & Original & if the man steps forward, will he fall onto the track \\
Answer & yes                            & Answer & no \\
Before & cars                           & Before & if the fall the \\
After & lights                          & After & if the steps forward, fall onto the \\\hline\\
                                         
& \includegraphics[height=1in]{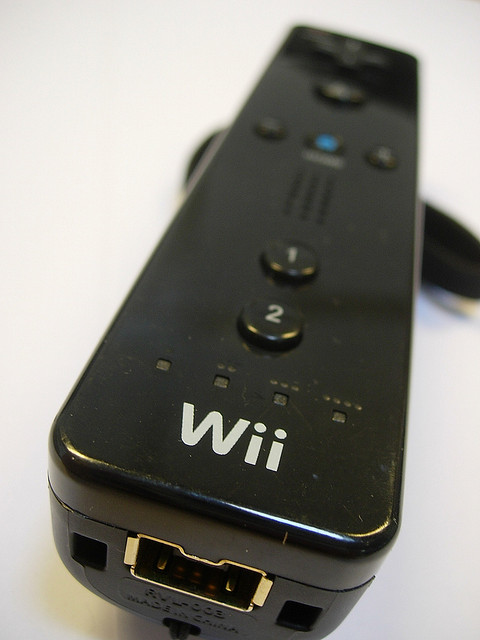}     & & \includegraphics[height=1in]{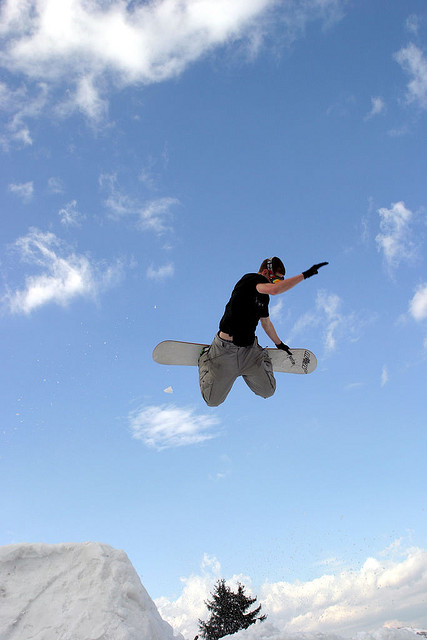}\\
Original & do this take batteries       & Original & are there any clouds in the sky \\
Answer & yes                            & Answer & yes \\
Before & batteries                      & Before & are clouds \\
After & batteries                       & After & there clouds sky \\\hline\\
                                         
& \includegraphics[height=1in]{126}     & & \includegraphics[height=1in]{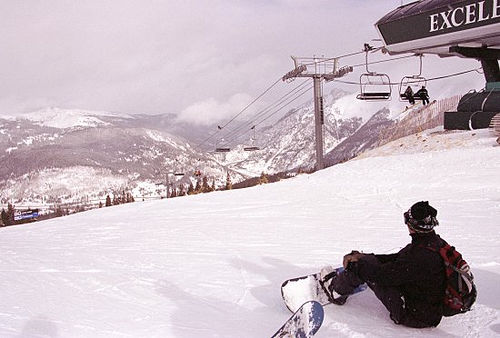}\\
Original & what is the man riding       & Original & what color is the skiers backpack \\
Answer & elephant                       & Answer & black \\
Before & man riding                     & Before & color \\
After & what riding                     & After & what color is the skiers \\\hline\\
                                         
& \includegraphics[height=1in]{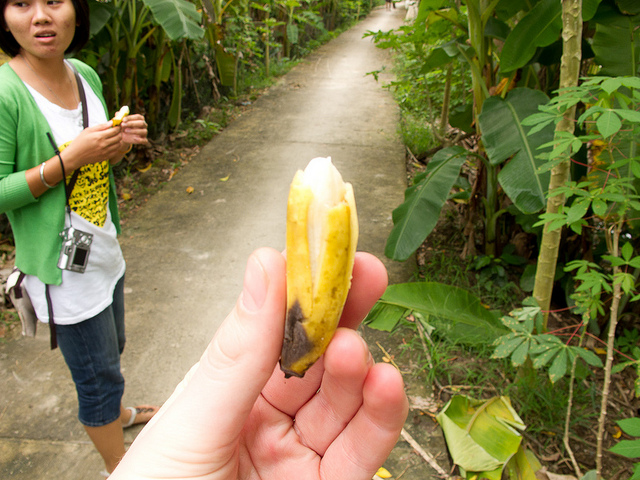}     & & \includegraphics[height=1in]{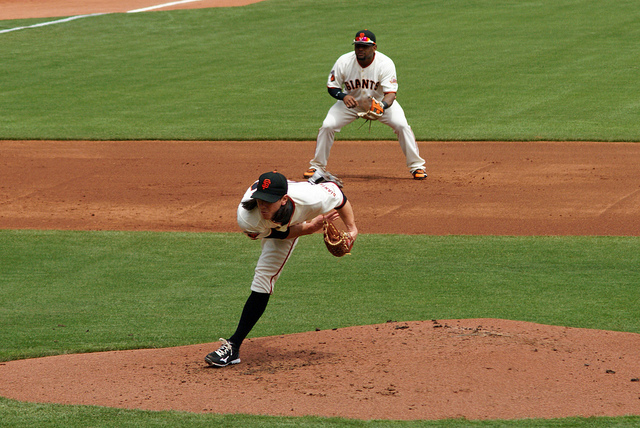}\\
Original & what kind of pants is the girl wearing & Original & are the giants playing \\
Answer & jeans                          & Answer & yes \\
Before & kind of pants wearing          & Before & are \\
After & what kind of pants              & After & giants playing \\\hline\\

\end{tabular}
\end{figure*}

\end{document}